\documentclass{article} 
\usepackage[preprint]{colm2025_conference}

\usepackage{amsmath,amsfonts,bm}

\def\eqref#1{equation~\ref{#1}}

\def\1{\bm{1}}

\DeclareMathAlphabet{\mathsfit}{\encodingdefault}{\sfdefault}{m}{sl}
\SetMathAlphabet{\mathsfit}{bold}{\encodingdefault}{\sfdefault}{bx}{n}

\DeclareMathOperator*{\argmax}{arg\,max}

\usepackage{microtype}
\usepackage{hyperref}
\usepackage{url}
\usepackage{booktabs}
\usepackage{lineno}

\usepackage[T1]{fontenc}

\usepackage{graphicx}
\usepackage{amsmath}
\usepackage{amssymb}
\usepackage{multirow}
\usepackage{subcaption}
\usepackage{colortbl}
\usepackage{enumitem}
\usepackage{fontawesome5}
\usepackage{wrapfig}
\usepackage[ruled,vlined]{algorithm2e}
\usepackage{caption}
\usepackage{csquotes}
\usepackage[breakable]{tcolorbox}

\definecolor{darkblue}{rgb}{0, 0, 0.5}
\hypersetup{colorlinks=true, citecolor=darkblue, linkcolor=darkblue, urlcolor=darkblue}

\newcommand\eg[0]{\textit{e.g.}}
\newcommand\ie[0]{\textit{i.e.}}

\title{\textit{Is Your LLM Secretly a World Model of the Internet}?\\ Model-Based Planning for Web Agents}

\author{
\quad\quad\quad\quad\quad\quad Yu Gu$^{1}$\thanks{Equal Contribution. See the contribution statement for details.}~, Kai Zhang$^{1}$\footnotemark[1]~, Yuting Ning$^{1}$\footnotemark[1]~, Boyuan Zheng$^{1}$\footnotemark[1]~, \\
\textbf{Boyu Gou$^{1}$, Tianci Xue$^{1}$, Cheng Chang$^{2}$, Sanjari Srivastava$^{2}$, Yanan Xie$^{2}$, Peng Qi$^{2}$,} \\
\quad\quad\quad\quad\quad\quad\quad\quad\quad\quad\quad\quad\quad\quad \textbf{Huan Sun$^{1}$\thanks{Equal Advising.}~, Yu Su$^{1}$\footnotemark[2]}
\\[1.2ex]
\quad\quad\quad\quad\quad\quad\quad\quad\quad 
\textsuperscript{1} The Ohio State University \quad \textsuperscript{2} Orby AI
\\
\quad\quad\quad\quad\quad\quad\quad\quad~~\small{\{gu.826, zhang.13253, sun.397, su.809\}@osu.edu}}

\newcommand{\OurMethod}{\textsc{WebDreamer}\xspace}

\begin{document}

\ifcolmsubmission
\linenumbers
\fi

\maketitle

\begin{abstract}
Language agents based on large language models (LLMs) have demonstrated great promise in automating web-based tasks.
Recent work has shown that incorporating advanced planning algorithms, \eg, tree search, is advantageous over reactive planning for web agents.
However, unlike simulated sandbox environments, real-world environments such as the web are rife with irreversible actions. 
This undermines the feasibility of backtracking, a cornerstone of (tree) search.  
Overly relying on test-time search also hurts efficiency. 
We advocate \textit{model-based planning} for web agents that employs a world model to simulate and deliberate over the outcome of each candidate action before committing to one. 
We systematically explore this paradigm by: \textbf{(1)} Proposing a model-based planning framework, \OurMethod, which employs LLMs to serve as both world models and value functions; \textbf{(2)} Training specialized LLMs as world models with a scalable data synthesis pipeline.
Empirical results demonstrate that \OurMethod achieves substantial performance improvements over reactive baselines.
It is competitive, while being $4$--$5$ times more efficient, with tree search in sandbox environments (VisualWebArena) and also works effectively on real-world websites (Online-Mind2Web and Mind2Web-Live).
Furthermore, our trained world model, Dreamer-7B, performs comparable to GPT-4o, highlighting the potential of specialized world models for efficient and effective planning in complex web environments.\footnote{All resources are available at \url{https://github.com/OSU-NLP-Group/WebDreamer}.}
\end{abstract}

\section{Introduction}
Planning~\citep{mattar2022planning}---deciding on optimal action sequences to achieve goals---has been fundamental to artificial intelligence since its inception.
Research into generalist web agents capable of planning and executing a sequence of actions to complete complex tasks across diverse websites has gained significant interest~\citep{mind2web,zhou2023webarena,zheng2024seeact,koh2024visualwebarena}, partly due to the web's potential as a complex yet realistic environment for driving agent research and development.
However, applying existing planning algorithms~\citep[\textit{inter alia}]{tot, world, gu-etal-2023-dont, wang2024planning, feng2023alphazero, brown2024large} to the online web environment presents formidable challenges.
Real-world environments such as the web are rife with state-changing and irreversible actions---for example, a single website like Amazon.com can involve numerous such actions, including submitting an order, creating an account, changing privacy settings, among many others---making \textit{backtracking}, a cornerstone of search-based planning~\citep{koh2024tree, putta2024agent}, highly challenging, if not infeasible.
The latency from excessive exploration in test-time search also hurts \textit{efficiency} and compromises user experience.

\begin{figure*}[tbh]
    \centering
    \includegraphics[width=\linewidth]{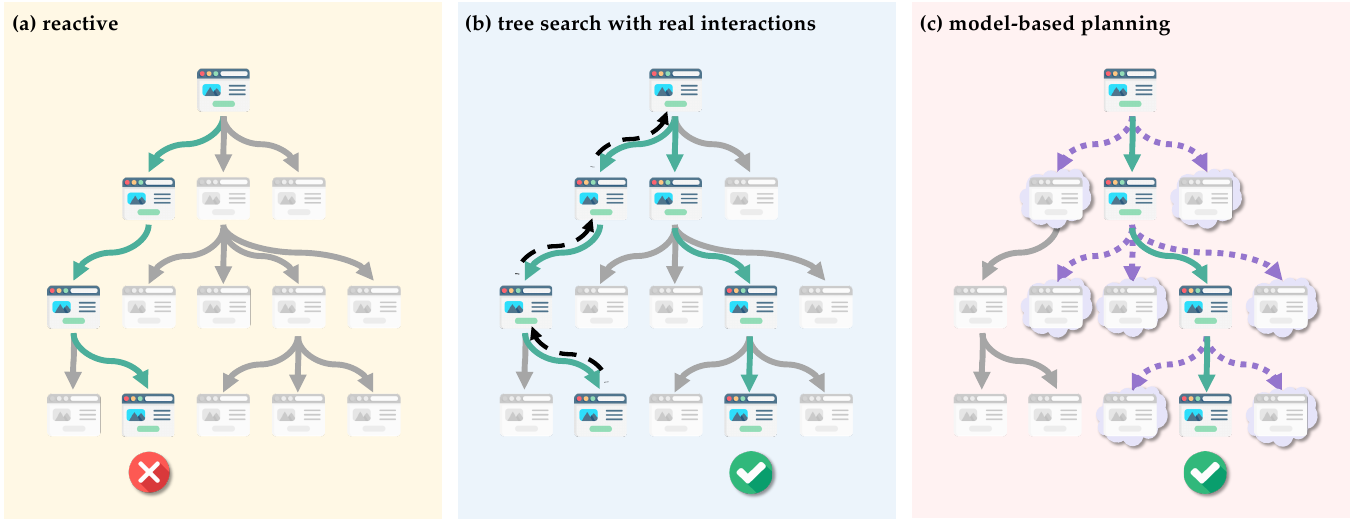}
    \captionof{figure}{Schematic illustration of different web agent strategies as a search problem, where each node represents a webpage.
    (a) Reactive: The agent selects locally optimal actions without planning, often leading to suboptimal results.
    (b) Tree search with real interactions: The agent explores multiple paths via active website navigation, potentially allowing backtracking (dashed arrows). However, backtracking is often infeasible due to irreversible actions.
    (c) Model-based planning: The agent simulates outcomes (cloud-bordered nodes) before execution, reducing real interactions while maintaining effectiveness.
    }
    \label{fig:overview}
\end{figure*}

One promising solution to address these challenges is \textit{model-based planning}~\citep{pascanu2017learning, moerland2023model}, which equips agents with the ability to simulate action sequences within a \textit{world model}---a computational representation of environment dynamics.
World models have achieved notable success~\citep{ha2018world, Hafner2020Dreamv1, hafner2021dreamerv2} in traditional reinforcement learning (RL) tasks within simulated environments like Atari games~\citep{Bellemare2013Atari, brockman2016openaigym}, where environment dynamics are well-defined and the action space is small and fixed, making world model training relatively straightforward.
However, building world models for web environments remains under-explored.
In contrast to the simulated environments, the Internet is open-ended and ever-evolving, with complex and diverse page structures and a wide range of possible user interactions.
This raises the question: \textit{How can we build effective world models for the Internet?}

We propose building world models for the Internet by leveraging large language models (LLMs) as the foundation.
Pretrained on web-scale data, LLMs have implicitly acquired both structural knowledge of websites and common sense needed to predict the outcomes of proposed actions, potentially making them well-suited to simulate transitions in complex web environments.
To this end, we introduce \OurMethod, a model-based planning framework that uses LLMs to simulate and score possible future states before executing actions, enabling informed decision-making (Figure~\ref{fig:overview}).
Building on this framework, we further train a dedicated world model, Dreamer-7B, using over 3.1 million interaction instances synthesized by our scalable data synthesis pipeline.

Empirical results demonstrate the effectiveness of our approach:
\OurMethod, when powered by state-of-the-art LLMs such as GPT-4o, achieves significant performance improvements over reactive baselines across three benchmarks, covering both online and sandbox environments.
It is also competitive with the tree search method, while being 4–5 times more efficient on sandbox environment VisualWebArena, and performs effectively on real-world websites (Online-Mind2Web and Mind2Web-Live), where tree search methods are difficult to implement and deploy.
Moreover, Dreamer-7B achieves performance comparable to GPT-4o on two online benchmarks.
In VisualWebArena, we can continue fine-tuning our Dreamer-7B with in-domain data synthesized by our pipeline.
With just 25K training instances, the resulting domain-specific world models yield even stronger results, surpassing GPT-4o.
These findings not only demonstrate the potential of using LLMs for model-based planning, but also establish a practical foundation for building world models for the open web through data synthesis, training, and evaluation.

\section{Related Work}
\subsection{Web Agents}
Web agents~\citep{Su2024LanguageAgents} powered by (multimodal) language models aim to automate web-based tasks, with benchmarks evolving from MiniWoB++~\citep{pmlr-v70-shi17a, liu2018reinforcement} to WebShop~\citep{yao2022webshop}, Mind2Web~\citep{mind2web}, WebArena~\citep{zhou2023webarena,koh2024visualwebarena}, which introduce more realistic environments and visual challenges. \textbf{Reactive Agents} make decisions based on immediate observations without simulation or search~\citep{react}. Enhancements include prompting proprietary models~\citep{zheng2024seeact, he2024webvoyager, mind2web} and training models on HTML and webpage screenshots~\citep{lee2023pix2struct, gur2023real, furuta2023multimodal, hong2024cogagent, baechler2024screenai}. 
Grounding improvements come from action-coordinate training~\citep{you2024ferret, cheng2024seeclick, gou2024uground}, while human-annotated~\citep{shaw2023pixels, hong2024cogagent, mind2web, lai2024autowebglm} and synthetic exploration trajectories~\citep{furuta2023multimodal, song2024trial, patel2024large, pahuja2025explorer} further refine agent behavior. However, these agents struggle with multi-step decision-making due to short-sightedness.
\textbf{Agents with Tree Search} have been explored to enhance decision-making. GPT-4V-based reward modeling~\citep{pan2024autonomous} and tree search algorithms~\citep{koh2024tree, putta2024agent, zhang2024webpilot} enable multi-step planning, with variants such as best-first search~\citep{koh2024tree} and MCTS~\citep{putta2024agent, zhang2024webpilot}. Despite performance gains, search methods significantly increase inference time, face challenges in backtracking on real websites, and risk unintended state-altering actions.

\subsection{World Models}
World models, central to model-based reinforcement learning (RL;~\cite{moerland2023model, sutton1991dyna}), learn state transitions to improve sample efficiency~\citep{ha2018world} and support planning~\citep{pascanu2017learning, schrittwieser2020mastering}.
Unlike traditional world models in RL focusing on improving data efficiency in the agent learning process, LLM-based world models emphasize decision-making over simulation fidelity, leveraging broad world knowledge for planning~\citep{world, kim2024cognitive}.
Our work extends this line by exploring LLM-based world models in complex web environments.
A concurrent work~\citep{chae2024web} also explores augmenting web agents with LLM-simulated action outcomes, however, their focus is to use small-scale data to train in-domain world models, while ours centers on understanding the potential of this new paradigm using advanced LLMs such as GPT-4o~\citep{GPT-4o} and self-trained general world models.

\section{\OurMethod: Model-Based Planning for Web Agents}
\label{sec:method}

\subsection{Preliminary}
\label{sec:planning}
Web agents tasked with automating activities in live websites face vast and complex search spaces. Formally, each task, given an instruction $I$, can be formulated as a partially observable Markov decision process (POMDP): $(\mathcal{S}, \mathcal{A}, \mathcal{O}, T, R, \Omega)$, where $\mathcal{S}$ is the set of possible environment states, $\mathcal{O}$ is the set of observations available to the agent, and $\mathcal{A}$ represents actions such as clicking elements, entering text, or navigating URLs.
$T: \mathcal{S} \times \mathcal{A} \to \mathcal{S}$ is the state transition function, while $R$ is a binary reward indicating task completion.
Due to partial observability, the agent perceives only observations $o = \Omega(s)$.

Tree search-based planning with real interactions is costly and risks irreversible actions.
Model-based planning through simulation mitigates this by using a learned simulation function $\texttt{sim}(o, a)$ to predict outcomes before execution.
This enables online planning, where the agent iteratively selects actions based on simulated future trajectories.
A common approach is Model Predictive Control (MPC;~\cite{garcia1989model}), which simulates possible future states for each action over a finite horizon $H$, evaluates them using a scoring function $\texttt{score}(\tau)$, and executes the action with the highest score.
This process repeats after observing new states, allowing adaptive decision-making while avoiding unnecessary interactions.

\begin{figure}
    \centering
    \includegraphics[width=\linewidth]{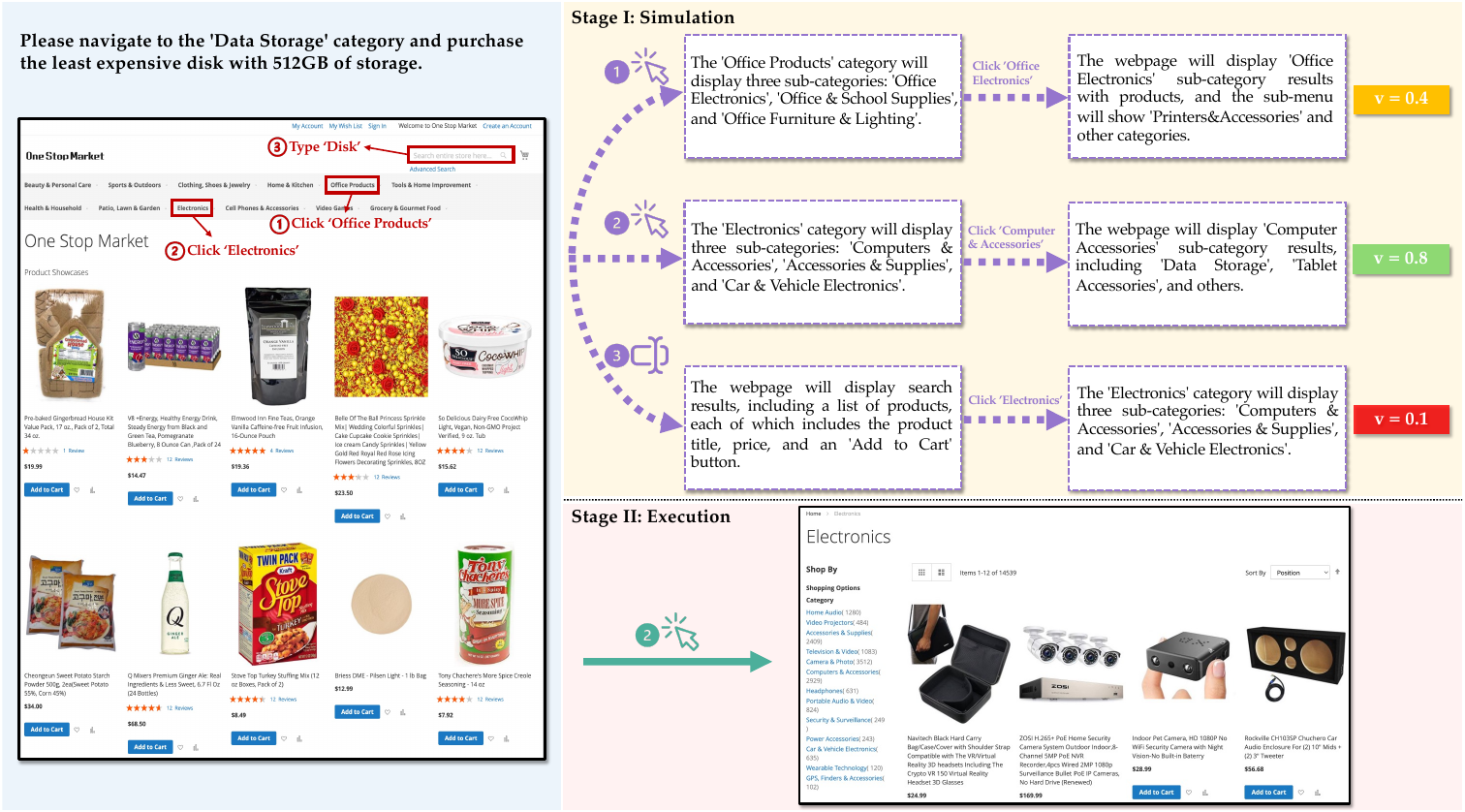}
    \caption{
     Illustration of \OurMethod simulating outcomes for three candidate actions using GPT-4o: (1) \textit{Click "Office Products"}, (2) \textit{Click "Electronics"}, and (3) \textit{Type "Disk" into textbox}.
     Each dotted box shows an LLM-generated state after a proposed action.
     Simulated trajectories are scored to identify the best action, and the optimal action \textit{Click "Electronics"} with the highest score ($v=0.8$) is executed.
     This example shows a two-step planning horizon.
     In practice, \OurMethod simulates multiple trajectories per action to capture a wider range of possible outcomes, improving coverage and leading to better-informed decisions.
     Here we only show one trajectory for each action and the final score for brevity.
    }
    \label{fig:diagram}
\end{figure}

\subsection{Core Design}
\label{subset:core-design}
\OurMethod follows the planning through simulation paradigm introduced in Section~\ref{sec:planning}. 
Figure~\ref{fig:diagram} illustrates this process with three candidate actions, where \OurMethod simulates two-step trajectories for each action, selects the trajectory with the highest score, and executes its corresponding initial action.
At its core, \OurMethod leverages LLMs to implement both the simulation function (\texttt{sim}) and the scoring function (\texttt{score}).

\begin{wrapfigure}{r}{0.45\textwidth}
\vspace{-1.5em}
    \centering
    \small
    \resizebox{0.45\textwidth}{!}{ 
        \begin{algorithm}[H]
            \caption{\OurMethod}
            \label{alg:planning}
            \SetAlgoLined
            \KwIn{Instruction $I$; initial observation $o_0$}
            \KwOut{Sequence of actions $a_0, a_1, \ldots, a_T$}
            $t\leftarrow 0$\;
            \While{True}{
                $\mathcal{A}_t\leftarrow \text{\texttt{get\_candidate}}(I, o_t)$\;
                $\mathcal{A}_t'\leftarrow \text{\texttt{self\_refine}}(\mathcal{A}_t)$\;
                $a_t = \argmax_{a \in \mathcal{A}_t'} \text{\texttt{score}}(\text{\texttt{sim}}(o_t, a))$\;
                $o_{t+1} \leftarrow \texttt{execute}(a_t)$\;
                $t \leftarrow t+1$\;
                \If{\texttt{termination\_check()} = True}{
                    break;
                }
            }
        \end{algorithm}
    }
    \vspace{-2em}
\end{wrapfigure}

\textbf{Implementation for \texttt{sim}.}
Our implementation of \texttt{sim} consists of two modules: one predicts state changes after action execution, approximating the state transition function $T$, while the other imagines a possible action based on the predicted state to enable long-horizon planning.
Together, these two modules generate trajectories of length $H$, where $H$ is a configurable parameter (the simulation depth).
Specifically, to represent the state changes, we prompt an LLM (GPT-4o or self-trained world models) to generate a concise natural language description focusing only on the effects of the action, as shown in Figure~\ref{fig:diagram} Stage I.

\textbf{Implementation for \texttt{score}.}
After collecting a trajectory $\tau_i$ simulated from each candidate action $a_i$ using \texttt{sim}, we further use an LLM as a scoring function for each simulation.
Following~\cite{koh2024tree}, we prompt GPT-4o to score each simulated trajectory with a three-scale response---complete (1.0), on track (0.5), or incorrect (0)---indicating its progress toward task completion. 
The final score for each action is averaged over multiple simulated trajectories and scorings, then used to determine the optimal action to execute (\eg, Click ``Electronics''), as shown in Stage I of Figure~\ref{fig:diagram}.

In addition to \texttt{sim} and \texttt{score}, a prerequisite to planning is candidate action generation.
We employ a two-stage approach: first sampling top-k actions following~\cite{koh2024tree}, then using LLM to self-refine unnecessary actions for simulation. 
This self-refinement step is motivated by our observation that at different steps, the same k can introduce varying degrees of irrelevant actions---some steps naturally have fewer plausible actions than others.
We show the pseudo code of \OurMethod's overall design in Algorithm~\ref{alg:planning}.
\texttt{termination\_check} verifies if the model outputs a \texttt{stop} action, reaches max steps, or repeats an action over 3 times, also following the implementation by~\cite{koh2024tree}. Please refer to Appendix~\ref{appendix:prompts} for more implementation details.

\subsection{World Model Data Synthesis and Model Training}
\label{subsec:train-world-model}

\begin{figure*}[tbh]
    \centering
    \includegraphics[width=\linewidth]{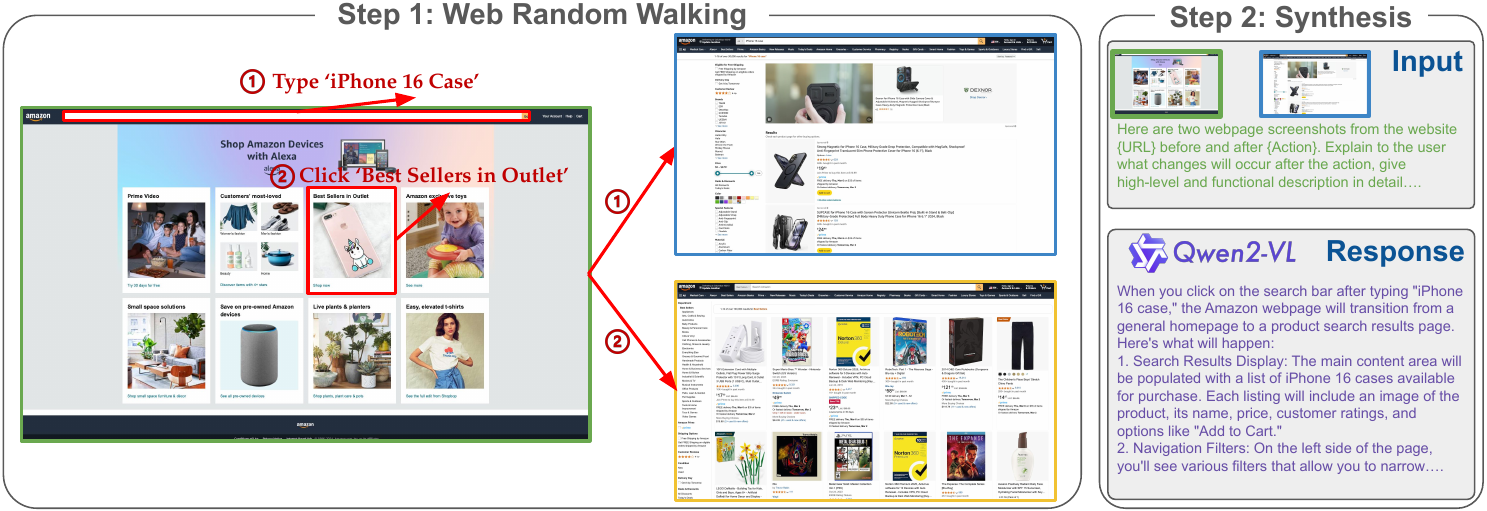}
    \captionof{figure}{Data synthesis pipeline. The pipeline consists of (1) Web Random Walking, where we autonomously interact with web pages through actions like typing for search and clicking, and (2) Synthesis, where Qwen2-VL-72B~\citep{Wang2024Qwen2VL} generates textual descriptions of state changes based on visual snapshots before and after actions.}
    \label{fig:data-construction}
\end{figure*}

While general-purpose LLMs such as GPT-4o have the potential to serve as world models, their cost and latency may limit their feasibility for real-time planning.
To explore a more deployable alternative, we also aim to train a small world model that offers lower inference cost and easier adaptation to new domains.

As shown in Figure~\ref{fig:data-construction}, we develop a scalable data synthesis pipeline that autonomously interacts with web pages using heuristic guidance.
We sample starting URLs from the October 2024 Common Crawl Index\footnote{https://commoncrawl.org} and perform random web actions, including clicking elements, hovering, typing into text boxes, and selecting options.
To better reflect human interaction distributions, we adjust action probabilities---favoring frequent interactions like clicking while ensuring sufficient coverage of others.
Additionally, we encourage causal dependencies by prioritizing actions on newly emerged elements, such as clicking a button that appears after hovering over a dropdown.
For search-related text inputs, we generate contextually relevant queries using GPT-3.5-turbo.

Once an interaction is performed, we capture visual snapshots before and after the action.
We then prompt Qwen2-VL-72B~\citep{Wang2024Qwen2VL} to generate textual descriptions detailing the changes in the webpage state (Figure~\ref{fig:data-construction} Step 2), ensuring an accurate representation of how each action impacts the visual content.
Each training instance consists of the initial visual state, the action taken, and the generated textual description of the state change.
After data collection, we filter out failed interactions, automation-blocked content, and potentially harmful data, resulting in a final dataset of over 3.1M interaction instances that capture rich causal relationships between user actions and web state transitions.

As we empirically find horizon step $H{=}1$ to be the most effective and efficient configuration, we focus on training the state transition function in \texttt{sim}, initializing it with Qwen2-VL-7B~\citep{Wang2024Qwen2VL}.
The final model, Dreamer-7B, is trained to predict the next state as a natural language description after performing an action on the current state, using a next-token prediction objective.
To efficiently monitor the progress of self-trained world models without relying on costly downstream evaluations for every checkpoint, we construct an intrinsic evaluation set for checkpoint selection, detailed in Appendix~\ref{appendix:intrinsic-eval}.
Appendix~\ref{appendix:data-synthesis-and-training} provides additional details on data synthesis and training.

\section{Experiments}
\label{sec:exp}
\subsection{Setup}

To properly test our planning framework's real-world performance, we focus on three representative web agent benchmarks, capturing the dynamic nature of web interactions: 
\noindent
\textbf{VisualWebArena} (VWA;~\cite{koh2024visualwebarena}) is designed to evaluate multimodal agents in visually grounded tasks. It includes 233 tasks verified by humans across three self-hosted websites: Classifieds, Shopping, and Reddit. The metric success rate is calculated as the percentage of tasks successfully accomplished by the agent-generated trajectories.
\noindent
\textbf{Online-Mind2Web}~\citep{m2wonline} is an online benchmark derived from Mind2Web~\citep{mind2web}, including 300 updated or newly created high-quality tasks spanning 136 real-world websites. 
These tasks can be categorized into easy, medium, and hard based on the number of steps required for completion. 
To reduce cost, we use a subset of 100 tasks, randomly sampling 30 easy, 40 medium, and 30 hard tasks.
The benchmark employs an automatic evaluation pipeline to measure task success rate, achieving an 85\% agreement with human judgment.
\noindent
\textbf{Mind2Web-Live}~\citep{webcanvas} consists of 104 tasks in 69 real-world websites refined from Mind2Web~\citep{mind2web}. 
It defines and annotates critical intermediate steps as key nodes for each task and considers a task successful only if all key nodes are completed.\footnote{We ensure no overlap between Online-Mind2Web and Mind2Web-Live for better task diversity.}
For all benchmarks, we use screenshots as the observation space and add Set-of-Mark~\citep{yang2023set} in VWA for fair comparison with the tree search baseline.
In our experiments, we empirically set the planning horizon $H$ to 1. A comprehensive analysis of this parameter is presented in Section~\ref{subsec:planning}.

To demonstrate the effectiveness of our framework and trained world models, we primarily compare our approach with two baselines: the reactive agent and the tree search agent with real interactions.\footnote{We will refer tree search with real interactions simply as tree search in our experiments for brevity.}
While we can readily implement our own method for all benchmarks, for the tree search baseline~\citep{koh2024tree}, we can only compare with it on VWA, due to the infeasibility of performing tree search on real-world websites in Online-Mind2Web or Mind2Web-Live.
Specifically, in VWA, \cite{koh2024tree} keep track of the sequences of actions to get to states in previous trajectories.
During backtracking, they reset the sandbox and re-execute the action sequence to restore the state. 
However, resetting the environment to undo effects is not feasible on real-world websites.

\subsection{Main Results}


\begin{table*}[bth]
    \centering
    \small
    \begin{tabular}{ccccc}
    \toprule
       Method &  World Model & VisualWebArena & Online-Mind2Web &  Mind2Web-Live \\
       \midrule
        Reactive & - & 17.6 & 26.0 & 20.2\\
        Tree Search  & - & \textbf{26.2} & - & - \\
        \midrule
        \multirow{4}{*}{\OurMethod} & GPT-4o & \underline{23.6} & \textbf{37.0} & \textbf{25.0} \\
         & Qwen2-VL-7B & 17.2 & 31.0 & 19.2 \\
         & Qwen2-VL-72B & 21.0 & 31.0 & 18.3\\
         & Dreamer-7B & 21.9 & \underline{35.0} & \underline{24.0} \\
    \bottomrule
    \end{tabular}
    \caption{Success rate (\%) on VisualWebArena~\citep{koh2024visualwebarena}, Online-Mind2Web~\citep{m2wonline}, and Mind2Web-Live~\citep{webcanvas}.
    We implement all the baselines ourselves to avoid discrepancies due to hardware and experimental settings in prior works.
    }
    \label{tab:downstream-eval}
    \vspace{-1em}
\end{table*}
\paragraph{Effectiveness.}
We present the overall performance results in Table~\ref{tab:downstream-eval}. 
\OurMethod demonstrates substantial improvements over the reactive agent on all benchmarks. 
Notably, on the VWA dataset, our proposed method achieves a 34.1\% relative performance gain and only trails behind tree search slightly.
It is important to note that tree search is not very practical on real-world websites, whereas \OurMethod provides a more flexible and adaptive alternative. On Online-Mind2Web and Mind2Web-Live, \OurMethod outperforms the reactive baseline by a relative gain of 42.3\% and 23.8\%, respectively. The strong results show the effectiveness of \OurMethod across different real-world websites.

Secondly, training world models on our large-scale synthesized data proves to be effective.
As shown in Table~\ref{tab:downstream-eval}, fine-tuning Qwen2-VL-7B into Dreamer-7B leads to a substantial 4.7\% absolute improvement in success rate on VisualWebArena, 4.0\% on Online-Mind2Web, and 4.8\% on Mind2Web-Live, outperforming the vanilla Qwen2-VL-7B model and even Qwen2-VL-72B.
Furthermore, Dreamer-7B achieves performance comparable to GPT-4o on two online benchmarks, Online-Mind2Web and Mind2Web-Live, demonstrating the feasibility of training world models for web-based decision-making.

\paragraph{Efficiency.}
Another key advantage of model-based planning is its efficiency compared with tree search using actual explorations.
Table~\ref{tab:efficiency} shows that tree search requires approximately 3 times more steps than the reactive baseline, whereas our method maintains comparable number of action steps.
Notably, compared to reactive baselines, tree search introduces about 10 times greater latency (in wall clock time) due to additional actions and backtracking, while the overhead from simulation in our approach is substantially lower, making \OurMethod 4--5 times more efficient than the tree search baseline.

\begin{table*}[h!]
\centering
\begin{minipage}{0.5\linewidth}
    \centering
    \resizebox{0.85\textwidth}{!}{ 
    \begin{tabular}{lccc}
        \toprule
        \textbf{Steps} & \textbf{Reactive} & \textbf{Tree Search} & \textbf{\OurMethod} \\
        \midrule
        Classifieds & 3.4 & 9.9 & 4.1 \\
        Reddit &  5.1 &  13.6 & 5.2  \\
        Shopping &  4.5 & 11.4 & 4.5  \\
        \bottomrule
    \end{tabular}
    }
    \caption*{(a) Number of action steps.}
\end{minipage}%
\hfill
\begin{minipage}{0.5\linewidth}
    \centering
    \resizebox{0.85\textwidth}{!}{ 
    \begin{tabular}{lccc}
        \toprule
        \textbf{Seconds} & \textbf{Reactive} & \textbf{Tree Search} & \textbf{\OurMethod} \\
        \midrule
        Classifieds & 68.3 & 749.2& 183.6  \\
        Reddit & 83.5 & 972.1 & 233.7    \\
        Shopping & 87.7 & 785.7  & 179.4  \\
        \bottomrule
    \end{tabular}
    }

    \caption*{(b) Task completion wall clock time.}
    
\end{minipage}
\caption{Efficiency analysis on VWA. All methods here use GPT-4o for fair comparison.}
\label{tab:efficiency}
\vspace{-1em}
\end{table*}

\section{Discussions}
\begin{wrapfigure}{r}{0.45\linewidth}
    \vspace{-1em}
    \centering
    \includegraphics[width=\linewidth]{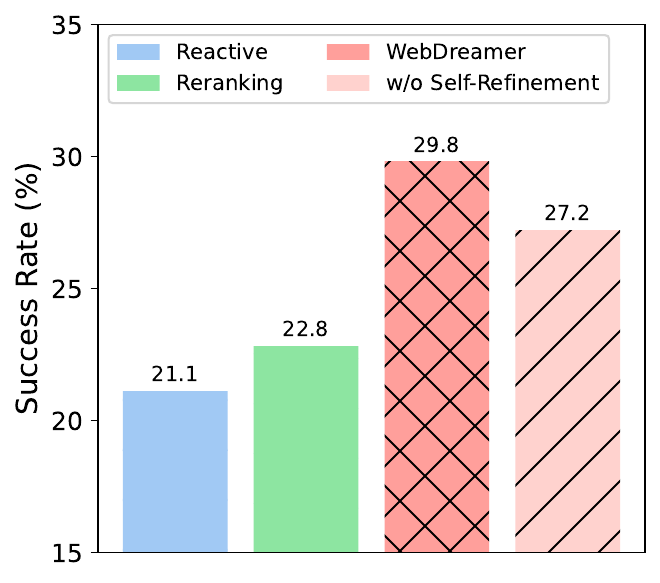}
    \vspace{-2em}
    \caption{Ablation study on the simulation stage and self-refinement stage.}
    \label{fig:ablation}
    \vspace{-1em}
\end{wrapfigure}

\subsection{Planning Framework}
\label{subsec:planning}
\paragraph{Ablation.}
We perform ablation studies on the simulation and self-refinement stages of \OurMethod on the VWA shopping human subset, which is the largest subset verified by humans.
We pay special attention to the simulation stage, which is the core of model-based planning.
One might argue that the primary improvement stems from reranking candidate actions, irrespective of whether this ranking relies on simulation.
To test this idea, we conduct an experiment where we remove the simulation stage completely and instead ask the reward model (\texttt{score}) to directly evaluate each candidate action (Reranking).
Additionally, we remove the self-refinement step after the action proposal in our \OurMethod framework to assess its impact (\textit{w/o} Self-Refinement).

As shown in Figure~\ref{fig:ablation}, this modified reranking approach does lead to some improvement over the reactive baseline, but the gain is small and still falls well behind \OurMethod. 
These results confirm that the LLM-based world model simulation plays a crucial role in the planning process.
Furthermore, we observe a decrease in performance when removing the self-refinement stage. 
Upon closer examination, we find that this decline is primarily due to the self-refinement module's ability to effectively filter out less relevant candidate actions when the next optimal action is clear. 
In contrast, directly simulating all actions may introduce additional noise that can negatively impact performance.

\paragraph{Planning Horizon.}
As introduced in Section~\ref{subset:core-design}, \OurMethod supports configurable planning horizon $H$ (\ie, the simulation depth).
To gain deeper insights into its effectiveness and current limitations, we investigate how the planning horizon affects the final performance. 
Using GPT-4o as the world model, we evaluate \OurMethod with planning horizons of 1, 2, and 3 on the same subset of Online-Mind2Web~\citep{m2wonline}.

\begin{figure}[t]
    \vspace{-0.5em}
    \centering
    \begin{minipage}{0.45\linewidth}
        \centering
        \includegraphics[width=\linewidth]{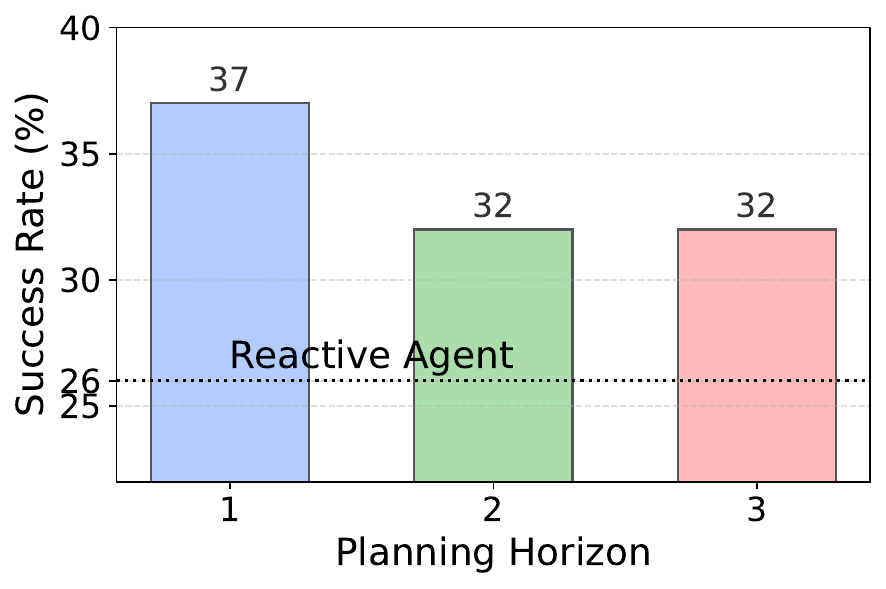}
        \vspace{-2.2em}
        \caption{Performance of \OurMethod with GPT-4o and different planning horizons $H$ on Online-Mind2Web.}
        \label{fig:planning_horizon}
        \vspace{-1em}
    \end{minipage}
    \hfill
    \begin{minipage}{0.45\linewidth}
        \centering
        \includegraphics[width=\linewidth]{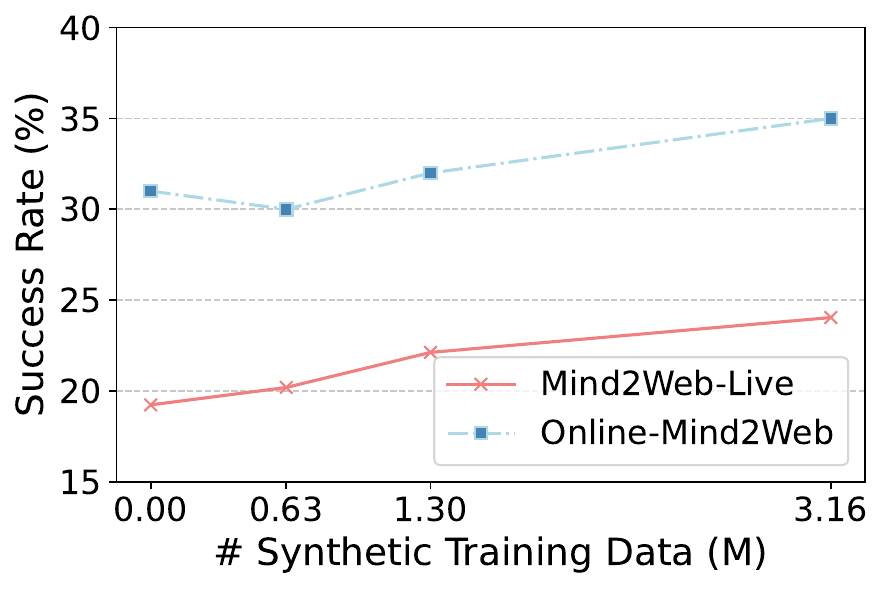}
        \vspace{-2.2em}
        \caption{Performance of \OurMethod with Dreamer-7B w.r.t. different sizes of training data on two online benchmarks.}
        \label{fig:scaling}
        \vspace{-1em}
    \end{minipage}
\end{figure}

As depicted in Figure~\ref{fig:planning_horizon}, the performance consistently outperforms the reactive baseline across all horizon settings. Nonetheless, when the planning horizon extends to 2 and 3 steps, the effectiveness begins to diminish.
Upon closer examination, this is primarily due to action proposal hallucinations within the simulation. Specifically, the action proposal within simulation is biased toward generating seemingly relevant actions for task completion, even when these actions are not available based on the predicted outcome.
As a result, as the planning horizon increases, the trajectories simulated from different actions become less distinguishable, as they all appear somewhat correct.
Given the complexity of the web environment, simulating multiple steps ahead is challenging due to error accumulation, which aligns with previous observations~\citep{mendes2025language, chae2024web}.
However, achieving better performance with longer horizons is not a major goal of this work; instead, we aim to show the feasibility and potential of using LLMs for model-based planning.
We leave this as a venue for future improvement.

\subsection{Training World Models}

\paragraph{Scaling Trend.}
We investigate the scaling trend by gradually increasing the size of our synthetic training data, as shown in Figure~\ref{fig:scaling}.
For Mind2Web-Live, performance steadily improves with larger training datasets, though the rate of improvement tapers off at higher data scales, suggesting potential diminishing returns.
In contrast, Online-Mind2Web shows more consistent gains overall, following a slight performance drop with small-scale training data.
These observations suggest that further scaling may continue to yield performance improvements. We omit scaling results on the sandbox environments in VWA and focus our analysis on the other two online benchmarks, which more accurately reflect real-world environments and web agent tasks.

\paragraph{In-Domain Continual Training.}
For specific environments, we can synthesize domain-specific data for world model training, enabling more specialized and contextually grounded simulations.
In VWA~\citep{koh2024visualwebarena}, we employ our data synthesis pipeline introduced in Section~\ref{subsec:train-world-model} to synthesize 25K in-domain interactions for each of the three environments: Classifieds, Reddit, and Shopping.
To prevent test data leakage, we filter out search actions containing queries that appear in test examples.
The final in-domain checkpoints are continually trained from the Dreamer-7B model, resulting in three separate world models, each specialized for its respective environment.

\begin{table}[htb]
\centering
\small
\begin{tabular}{clcccc}
\toprule
\multicolumn{1}{l}{}        &                & Classifieds   & Reddits       & Shopping    & Total      \\ \midrule
\multicolumn{2}{l}{Reactive}                 & 17.9          & 14.3          & 19.3        & 17.6  \\
\multicolumn{2}{l}{Tree Search~\citep{koh2024visualwebarena}}              & \textbf{26.8} & \textbf{20.6} & \textbf{28.9} & \textbf{26.2} \\ \midrule
\multirow{5}{*}{WebDreamer} & GPT-4o         & 23.2          & \underline{17.5}    & \underline{26.3}  & \underline{23.2}       \\
                            & Qwen2-VL-7B       & 17.9          & 11.1          & 20.2   & 17.2       \\
                            & Qwen2-VL-72B      & 19.6          & 15.9          & 24.6   & 21.0       \\
                            & Dreamer-7B & 21.4          & 15.9          & 25.4   & 21.9         \\
                            & ~~$+$ In-Domain   & \underline{25.0}    & 15.9          & \underline{26.3}  & \underline{23.2} 
                            \\ \bottomrule
\end{tabular}
\caption{Success rate (\%) of \OurMethod with various world models on VWA.}
\label{tab:in-domain}

\end{table}

As shown in Table~\ref{tab:in-domain}, continual training improves performance over the Dreamer-7B model and achieves results comparable to or even better than GPT-4o in certain environments.
The Classifieds and Shopping domains benefit the most from in-domain adaptation, demonstrating that domain-specific fine-tuning successfully refines model predictions to better reflect the specific environment dynamics.
However, performance on Reddit remains unchanged, likely due to its dense viewport and limited functional interactions.
Unlike Classifieds and Shopping, which have simpler visual organization, Reddit pages contain long, text-heavy content, making it difficult for a world model to infer meaningful state changes beyond surface-level text visibility.
In addition, the limited perception abilities~\citep{gou2024uground, zhang2025chartEval} of 7B models in dense web viewports may further constrain them to simulate fine-grained changes in text-heavy environments.
These results underscore the effectiveness of domain-specific adaptation while highlighting areas for further improvement in specific web domains.
Future work can explore enhanced representation techniques to better handle dense web layouts, further expanding the applicability of world models across diverse real-world web environments.

\paragraph{Case Studies.}

\begin{figure*}[tbh]
    \centering
    \includegraphics[width=\linewidth]{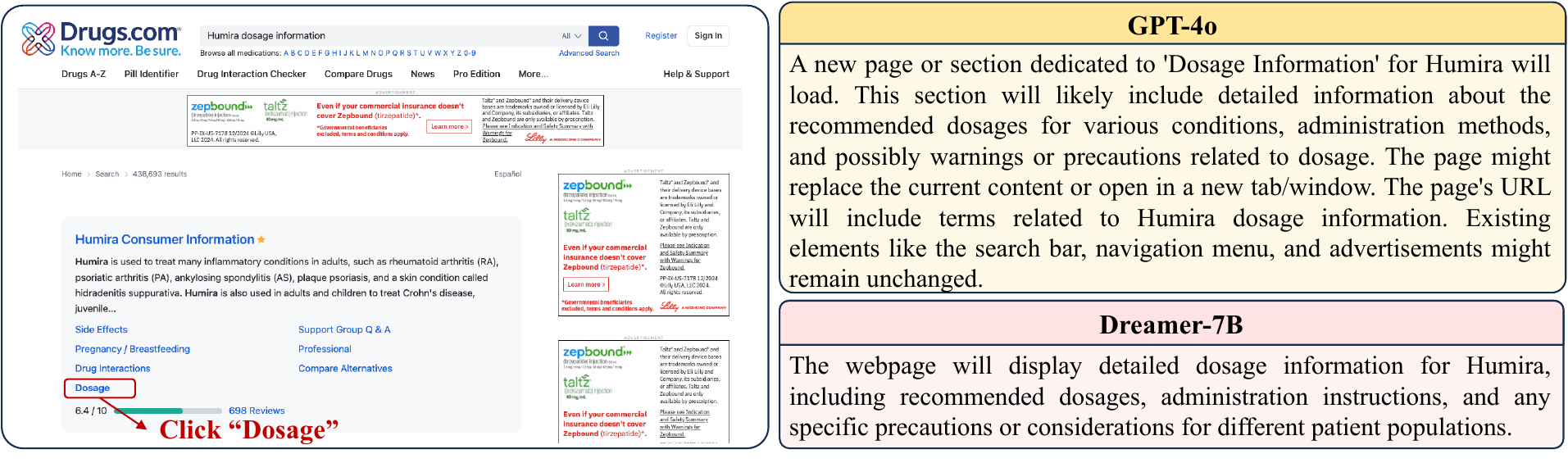}
    \vspace{-1em}
    \captionof{figure}{An example comparing GPT-4o and Dreamer-7B as world models. General-purpose models such as GPT-4o often provide detailed but sometimes noisy predictions, while our Dreamer-7B focuses on concise, action-relevant outcomes.}
    \label{fig:case-study}
\end{figure*}

We present a case study to explore the differences between GPT-4o and Dreamer-7B as world models.
GPT-4o usually provides more detailed state change descriptions and considers multiple possible outcomes, though some of these details are unimportant or irrelevant. 
For example, as shown in Figure~\ref{fig:case-study}, GPT-4o successfully predicts the key state changes after clicking on ``Dosage,'' outlining the relevant information that may appear on the new webpage. 
However, its prediction also includes unimportant details such as URL and UI changes, which may introduce noise for the policy model. 
On the contrary, Dreamer-7B offers more concise and action-oriented predictions, directly predicting the most important information on the new webpage.

To further clarify the role of simulation in planning, we also present case studies with GPT-4o as a world model, covering both positive and negative examples in Appendix~\ref{appendix:case-study}.
They illustrate how simulation aids the agent in exploring the environment, as well as how inaccuracies in simulation can lead to incorrect predictions.

\section{Conclusion}
In this paper, we systematically explore the use of LLMs as world models for model-based planning in web environments.
Our planning framework, \OurMethod, achieves substantial improvements over reactive baselines across three benchmarks and offers a $4$--$5$ times more efficient yet competitive alternative to tree search, which is often impractical due to backtracking and efficiency constraints on real-world websites.
Beyond using off-the-shelf LLMs as world models, we train Dreamer-7B on 3.1 million web interaction examples synthesized by our scalable data synthesis pipeline, achieving performance comparable to GPT-4o.
Furthermore, continual in-domain fine-tuning enables Dreamer-7B to adapt to specific environments, further improving simulation quality and downstream performance---surpassing GPT-4o.
This work lays the foundation for future research on model-based planning methods for efficient and effective decision-making in web environments.
It also establishes groundwork for further scaling world models to handle general web tasks.

\section*{Acknowledgments} 
We would like to extend our appreciation to colleagues from the OSU NLP group and Orby AI for their insightful comments. 
This work is supported in part by Orby AI and ARL W911NF2220144. 
The views and conclusions contained herein are those of the authors and should not be interpreted as representing the official policies, either expressed or implied, of the U.S.\ government. The U.S.\ government is authorized to reproduce and distribute reprints for government purposes notwithstanding any copyright notice herein.

\bibliography{colm2025_conference}
\bibliographystyle{colm2025_conference}

\newpage
\appendix

\counterwithin{figure}{section}  
\counterwithin{table}{section}   

\renewcommand{\thefigure}{\thesection.\arabic{figure}}
\renewcommand{\thetable}{\thesection.\arabic{table}}

\section*{Overview}

The appendix includes the following sections:
\begin{itemize}
    \item \textbf{Section~\ref{appendix:contribution-statement}: Author Contribution Statement.}
    \item \textbf{Section~\ref{appendix:prompts}: Implementation Details of \OurMethod.}
    \item \textbf{Section~\ref{appendix:data-synthesis-and-training}: Implementation Details of Data Synthesis and World Model Training.}
    \item \textbf{Section~\ref{appendix:intrinsic-eval}: Intrinsic Evaluation of World Models.} 
    \item \textbf{Section~\ref{appendix:case-study}: Case Study.}
\end{itemize}

\section{Contribution Statement}
\label{appendix:contribution-statement}
\textbf{Yu Gu} conceived the project with Yu Su and developed the planning framework \OurMethod.
He implemented the core codebase, led the design and execution of key experiments including ablation studies, and wrote the initial manuscript draft.
He also led the public release of the resources.

\textbf{Kai Zhang} led the data synthesis and world model training efforts, including the training of Dreamer-7B and domain-specific variants.
He conducted experiments on VisualWebArena and played a major role in editing and polishing the manuscript to its final form.

\textbf{Yuting Ning} led the development of the intrinsic evaluation suite and conducted experiments on Online-Mind2Web and Mind2Web-Live.
She also contributed significantly to the final manuscript refinement.

\textbf{Boyuan Zheng} contributed to the development of the planning framework, the data synthesis pipeline, and the in-domain world model training. He also provided valuable insights into the experimental design.

\textbf{Orby AI team} (Cheng Chang, Sanjari Srivastava, Yanan Xie, and Peng Qi) provided large-scale compute resources and data crawling support, which were essential for scaling up the data synthesis and training pipeline. Peng Qi and Yanan Xie also provided constructive feedback throughput the project.

\textbf{Yu Su} and \textbf{Huan Sun} steered the main directions throughout the project, led discussions on the planning framework and experiments, and provided funding support.
\textbf{Yu Su} conceived the project with Yu Gu.

All authors reviewed the manuscript and provided feedback.

\section{Prompts for Four Stages in \OurMethod}
\label{appendix:prompts}
\subsection{Action Proposal}
\label{appendix:action-proposal}
\begin{tcolorbox}[colframe=green!50!black, colback=green!10!white, title=Action Proposal, breakable]
\small 
You are an autonomous intelligent agent tasked with navigating a web browser. You will be given web-based tasks. These tasks will be accomplished through the use of specific actions you can issue.

Here's the information you'll have: \{\texttt{Web Information}\}

The user's objective: \{\texttt{Task Objective}\} This is the task you're trying to complete.

The current web page screenshot: \{\texttt{Web Page Screenshot Image}\} This is a screenshot of the webpage, with each interactable element assigned a unique numerical id. Each bounding box and its respective id shares the same color.

The observation, which lists the IDs of all interactable elements on the current web page with their text content if any, in the format \texttt{[id][tagType][text content]}. \texttt{tagType} is the type of the element, such as button, link, or textbox. \texttt{text content} is the text content of the element. For example, \texttt{[1234][button][`Add to Cart']} means that there is a button with id 1234 and text content \texttt{`Add to Cart'} on the current web page. \texttt{[][StaticText][text]} means that the element is of some text that is not interactable.

The current web page's URL: \{\texttt{Web URL}\} This is the page you're currently navigating.

The open tabs: \{\texttt{Previous Tabs}\} These are the tabs you have open.

The previous action: \{\texttt{Previous Action}\} This is the action you just performed. It may be helpful to track your progress.

The actions you can perform fall into several categories:

\smallskip
Page Operation Actions:

~~~- \texttt{click [id]}: This action clicks on an element with a specific id on the webpage.

~~- \texttt{type [id] [content]}: Use this to type the content into the field with id.
By default, the \texttt{Enter} key is pressed after typing unless \texttt{press\_enter\_after} is set to 0, \ie, \texttt{type [id] [content] [0]}.

~~- \texttt{hover [id]}: Hover over an element with id.

~~- \texttt{press [key\_comb]}: Simulates the pressing of a key combination on the keyboard (\eg, Ctrl+V)

~~- \texttt{scroll [down]} or \texttt{scroll [up]}: Scroll the page up or down.

\medskip

Tab Management Actions:

~~- \texttt{new\_tab}: Open a new, empty browser tab.

~~- \texttt{tab\_focus [tab\_index]}: Switch the browser's focus to a specific tab using its index.

~~- \texttt{close\_tab}: Close the currently active tab.

\medskip

URL Navigation Actions:

~~- \texttt{goto [url]}: Navigate to a specific URL.

~~- \texttt{go\_back}: Navigate to the previously viewed page.

~~- \texttt{go\_forward}: Navigate to the next page (if a previous \texttt{go\_back} action was performed).

\medskip

Completion Action:

~~- \texttt{stop [answer]}: Issue this action when you believe the task is complete. If the objective is to find a text-based answer, provide the answer in the bracket.

\medskip

Homepage:

If you want to visit other websites, check out the homepage at \underline{http://homepage.com}. It has a list of websites you can visit.
\underline{http://homepage.com/password.html} lists all the account name and password for the websites. You can use them to log in to the websites.

\medskip

To be successful, it is very important to follow the following rules:

~~1. You should only issue an action that is valid given the current observation

~~2. You should only issue one action at a time.

~~3. You should follow the examples to reason step by step and then issue the next action.

~~4. Generate the action in the correct format. Start with a \textit{``In summary, the next action I will perform is''} phrase, followed by action. For example, \textit{In summary, the next action I will perform is }\texttt{click [1234]}.

~~5. Issue stop action when you think you have achieved the objective. Don't generate anything after stop.

\end{tcolorbox}

\subsection{Self-Refinement}
\label{appendix:controller}

\begin{tcolorbox}[colframe=blue!50!black, colback=blue!10!white, title=Self-Refinement, breakable]
\small
You are assisting a web navigation agent to help a human user navigate a website to complete a task. Given the user's intent, the action history, and the current state of the webpage, the agent has proposed a set of candidate actions to take at the current step. 

\medskip
Your role is not to determine a best action for the agent at this step, but to filter out the actions that are very likely not relevant or helpful for the agent to accomplish the task.

\medskip
Please select all actions that you think that could possibly lead the agent to accomplish the task. It's important to note that to accomplish a task, the agent will execute a sequence of actions. So the action to take at this step does not have to immediately lead to the completion of the task. You should select any action that could be relevant for the agent to take in the current state of the webpage. Try to be as thoughtful and comprehensive as you can! Don't miss any possible action. If there is one action that is clearly the best, and all other actions are clearly not very relevant, you can only select one action. Please do this sparely, since some actions may be helpful in a longer horizon. 

\medskip
An action should be included as long as it could be relevant to the task, even if it may not be the most direct action to take at this step!! Some relevant actions might seem indirect at the first glance, but could be helpful in a longer horizon. Please also include those actions.

\medskip
Please at least select one action.

\bigskip
\textbf{*IMPORTANT*}

Format your response into two lines as shown below:

\bigskip
Thoughts: \texttt{<your thoughts and reasoning process>}. You must explicitly evaluate each action one by one and imagine whether it could be relevant to the task following the format: \texttt{action:... rationale:...}

\medskip
Selected actions: \texttt{id0;id1;aid2;...}
(please return the index of the action in the candidate actions list, starting from 0. Don't output the action description itself. Separate the indices with semicolons. Do not add spaces or any other characters after the semicolons.)

\bigskip
Action History: \{\texttt{last\_actions\_str}\}

\bigskip
Current URL: \{\texttt{current\_url}\}

\bigskip
The images corresponding to the user intent are shown in the FIRST \{\texttt{len(intent\_images)}\} images (before the User Intent).

\bigskip
The last \{\texttt{len(screenshots)}\} snapshots of the agent's trajectory are shown in the LAST \{\texttt{len(screenshots)}\} images. The LAST IMAGE represents the current state of the webpage.

\bigskip
Proposed Action: \{\texttt{action\_descriptions}\}

\end{tcolorbox}

\subsection{World Model}
\label{appendix:world-model}

\begin{tcolorbox}[colframe=red!50!black, colback=red!10!white, title=World Model (state transition function), breakable]
\small

You are an agent that predicts the effect of an action on a webpage. You will be given a screenshot of a webpage, a sequence of actions and state changes applied to the initial screenshot, and an operation to perform on the webpage. You are required to predict the new changes that will occur on the webpage after the operation is performed, such as the appearance of new elements, the disappearance of existing elements, or changes in the content of existing elements. The operation type and the element to operate will be provided in the prompt. Directly output~~\texttt{State changes:...} and don't output anything else. Try to be as comprehensive and detailed as possible.

\medskip
Based on the initial screenshot and the changes to the webpage, please predict the changes after action: 

\end{tcolorbox}

\subsection{Reward Model}
\label{appendix:reward-model}

\begin{tcolorbox}[colframe=yellow!50!black, colback=yellow!10!white, title=Reward Model (\texttt{score}), breakable]
\small
You are an expert in evaluating the performance of a web navigation agent. The agent is designed to help a human user navigate a website to complete a task. Given the user's intent, the agent's action history, the current state of the webpage, your goal is to decide \textbf{**whether the simulated steps by the agent indicate a successful execution of the user intent**}. In particular, if the predicted state (\ie, the current state represented by the last image plus all the predicted changes so far) corresponds to a successful final state. If it is a failure but it looks like the simulated steps are on the right track towards success, you should also output as such. Note that, in the simulated steps, all the state changes are predicted by the agent's world model, and they may not actually be faithful to the real website interactions (\eg, some proposed actions may not be available in a realistic website). You should also account for this in your evaluation (\eg, if the predicted state changes are not reasonable then it's probably a failure).

\medskip
\textbf{*IMPORTANT*}

\medskip
Format your response into two lines as shown below:

\medskip
Thoughts: \texttt{<your thoughts and reasoning process>}

\smallskip
Status: \texttt{``success''} or \texttt{``failure''}

\smallskip
On the right track to success: \texttt{``yes''} or \texttt{``no''}

\end{tcolorbox}

\section{Data Synthesis and World Model Training}

\label{appendix:data-synthesis-and-training}
\subsection{Data Synthesis Prompt}

\begin{table}[tbh]
\centering
\small
\begin{tabular}{p{0.90\linewidth}}
\toprule
\textbf{Prompt Templates} \\
\midrule
Here is the web screenshot <image\_token>. Please describe what you would see after performing \{\texttt{action}\} on \{\texttt{element\_description}\}. \\ \midrule
Here is the web page you are looking at <image\_token>. Please describe what you would see after doing \{\texttt{action}\} on \{\texttt{element\_description}\}. \\ \midrule
Here is the web page you are currently at <image\_token>. Describe what you will see after \{\texttt{action}\}ing \{\texttt{element\_description}\}. \\ \midrule
Here is the current web page <image\_token>. Briefly describe what you will see after \{\texttt{action}\}ing \{\texttt{element\_description}\}. \\ \midrule
Below is the current screenshot <image\_token>. Briefly describe what you will see after \{\texttt{action}\}ing \{\texttt{element\_description}\}. \\ \midrule
Below is the current screenshot <image\_token>. Describe what you will see after \{\texttt{action}\}ing \{\texttt{element\_description}\}. \\ \midrule
Below is the current screenshot <image\_token>. Please describe what you would see after a \{\texttt{action}\} on \{\texttt{element\_description}\}. \\
\bottomrule
\end{tabular}
\caption{Prompt templates used to generate language descriptions of state transitions.}
\label{tab:prompt-templates}
\end{table}

In practice, we first draw a red bounding box around the target element to precisely localize it for Qwen2-VL-72B.
Next, we prompt Qwen2-VL-72B to separately describe the element using a referring expression and to describe the resulting state change.
Finally, we combine the referring expression and the state change description to construct the training instance, using one of the templates randomly selected from Table~\ref{tab:prompt-templates}.
We list the prompt used to synthesize natural language descriptions of the next state below.

Despite the fact that our training data includes only a few prompt templates and natural images, experiments in Section~\ref{sec:exp} have shown that the model generalizes well to unseen instruction or prompts used in benchmarks like Online-Mind2Web~\citep{m2wonline} and Mind2Web-Live~\citep{webcanvas} and to images with Set-of-Mark~\citep{yang2023set} in VisualWebArena~\citep{koh2024visualwebarena}.

\newpage
\begin{tcolorbox}[colframe=white!50!black, colback=white!10!white, title=Data Synthesis, breakable]
\small
\label{box:data-synthesis-prompt}

Here are two webpage screenshots from the website \{URL\} before and after \{Action\} on the element within the red bounding box.

\medskip
**\textbf{Element Description:}** Please describe the element within the bounding box to ensure user can locate this element in the webpage image only use this description (only element description, not bounding box).

So DO NOT say something like the element within the bounding box, DO NOT SAY anything about red bounding box.

Instead, describe the element with referral expression like the button showing the text `Make Appointment' or the `Search' in the search bar.

Starts with lower case and make sure the description is a noun like the element that is a category link labeled `Massage' located in the sidebar on the left side of the page.

\medskip
**\textbf{Change Description:}** Explain to the user what changes will occur on the webpage after they click on the described element. Do not say too much trivial details.
Instead, give high-level and functional description in detail after the action.
Focus solely on describing the changes that will happen, not the element.


\end{tcolorbox}

\subsection{Training Data Statistics}
\begin{table}[tbh]
\centering
\small
\begin{tabular}{lcc}
\toprule
            & \multicolumn{1}{c}{Number} & \multicolumn{1}{c}{Percentage} \\ \midrule
Unique URLs & 1,247,960                    & \multicolumn{1}{c}{-}         \\ \midrule
Action      & \multicolumn{1}{c}{}       & \multicolumn{1}{c}{}          \\
~~- Click       & 2,653,704                    & 84.0                          \\
~~- Hover       & 241,234                     & 7.6                           \\
~~- Type        & 217,692                     & 6.9                           \\
~~- Select      & 47,617                      & 1.5                           \\
Total       & 3,160,247                    & 100                          \\ \bottomrule
\end{tabular}
\caption{Training data distribution.}
\label{tab:appendix-training-data-dist}
\end{table}

As described in Section~\ref{subsec:train-world-model}, we develop a scalable pipeline to synthesize large-scale interaction data by randomly exploring web pages across a wide range of domains.
This process results in over \textbf{3.1 million} interaction instances spanning \textbf{1.2 million unique URLs}, as summarized in Table~\ref{tab:appendix-training-data-dist}.
We focus on four primary interaction types: \textit{click}, \textit{hover}, \textit{type}, and \textit{select}, which collectively reflect the core user intents on modern web interfaces. Clicks dominate the dataset (84\%), consistent with their central role in triggering state transitions on the web.
Although our data synthesis pipeline initially included \textit{scroll} actions, we empirically found that they contributed little to downstream performance and thus excluded them from the final training set.

\begin{figure}
\centering
\includegraphics[width=0.8\linewidth]{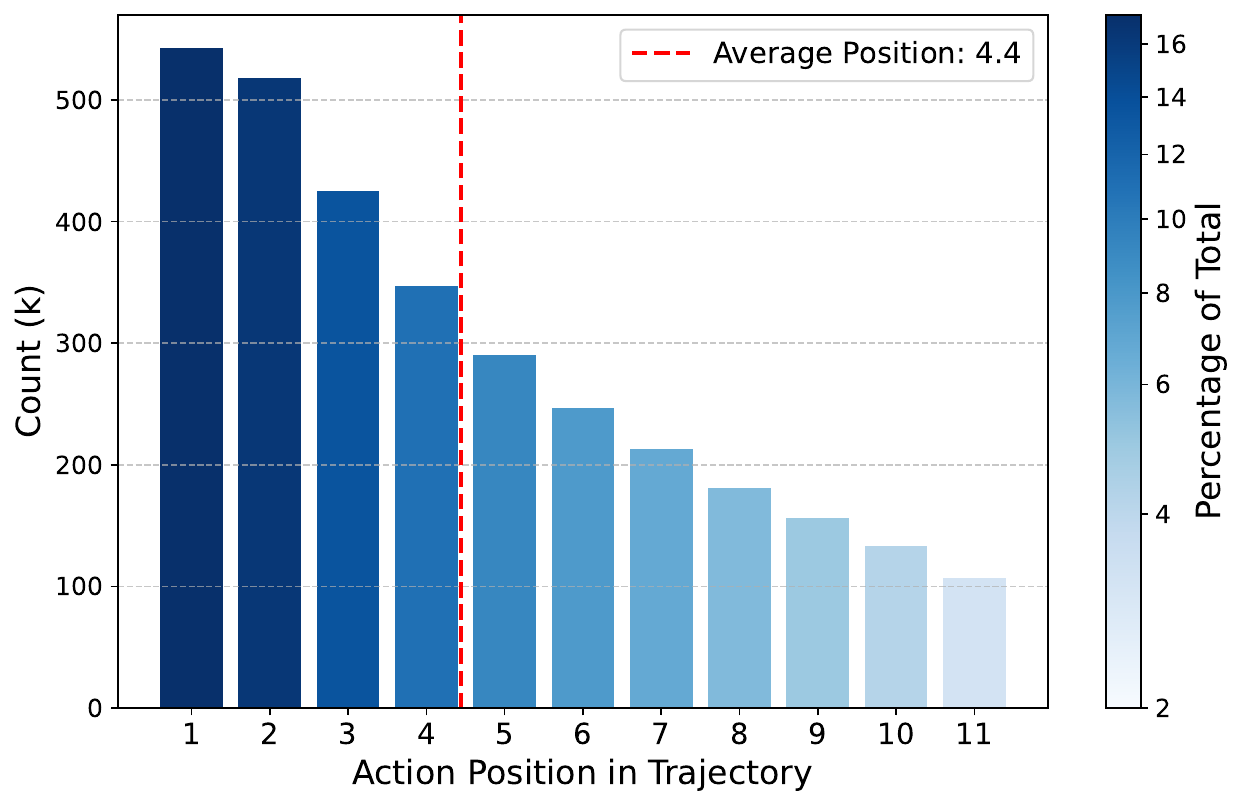}
\caption{Distribution of the step at which actions occur within multi-step trajectories. The actions used for training span across different depths, not being limited to initial steps.}
\label{fig:action-position-stat}
\end{figure}

Importantly, the interactions in our dataset are not restricted to initial steps but span various stages within multi-step trajectories.
Figure~\ref{fig:action-position-stat} shows the distribution of actions by their position in the interaction sequence.
The average position is 4.4, and a substantial portion of actions occur deeper in the trajectory.
This distribution allows the world model to learn more informative web dynamics and have the potential to enable long-term planning.

\subsection{World Model Training}

All world models are fine-tuned using the Qwen2-VL-7B-Instruct~\citep{Wang2024Qwen2VL} backbone.
To align with future state prediction, we format training examples using structured prompts, such as:
\textit{``Here is the web screenshot. Please describe what you would see after performing \{action\} on \{element\}."}
All experiments are conducted on 64 H100 GPUs with 80GB memory each.
The final Dreamer-7B model is trained for up to 2 epochs over the full training dataset described above.
We evaluate models every 1000 steps using an intrinsic evaluation metric (described later) to ensure fair and consistent model selection across settings.
We use the DecoupledAdamW~\citep{loshchilov2018decoupled} optimizer with a learning rate of \texttt{1e-6}, $\beta_1 = 0.9$, and $\beta_2 = 0.95$.
The learning rate follows a cosine decay schedule with 1000 warmup steps, decaying to 10\% of the base LR.
We use a global batch size of 192 and enable mixed-precision training.

For the three domain-specific world models (Classifieds, Reddit, and Shopping), we continue training from the Dreamer-7B checkpoint for one epoch using the respective in-domain datasets.
These models follow the same training setup, except with a reduced learning rate of \texttt{5e-7} and a shorter warmup of 100 steps to ensure stable adaptation on smaller datasets.



\section{Intrinsic Evaluation of World Models}

\label{appendix:intrinsic-eval}

While downstream evaluation with real-world web tasks can provide valuable insights for the model-based planning framework, it involves many other factors (\eg, the policy model) that may influence performance other than world models.
Therefore, to rigorously evaluate world models only, we construct an intrinsic evaluation that provides a more controlled and independent assessment while maintaining alignment with the downstream evaluation setting.
When training our world models, we use this intrinsic evaluation for model development and checkpoint selection.

\subsection{Dataset Construction}

We first sample tasks from Mind2Web~\citep{mind2web} and manually annotate the trajectories that can complete the tasks as ground truths. Then, we generate deviation trajectories at different states in the ground truth trajectories and assess whether the world model can help distinguish the correct actions from incorrect actions at each state. Specifically, for each state on the ground truth trajectory, we use the ranking model in MindAct~\citep{mind2web} to select top-5 candidate elements from the webpage that align best with the task intention and current state, excluding the ground truth one. These deviation actions are then extended into full trajectories using a web agent (\ie, SeeAct~\citep{zheng2024seeact}), automatically evaluated with an LLM-as-a-judge method~\citep{pan2024autonomous}.
Since multiple paths may lead to task completion, we filter out deviation actions on successful trajectories to alleviate false negative issue and only retain those in failed trajectories as negative actions. Using this pipeline, we construct an intrinsic evaluation dataset of 44 tasks, 141 states with deviations, and 279 deviation actions.

\subsection{Evaluation Metrics}

For evaluation, we use world models to simulate possible future states for each action and score them using the same scoring function in the planning framework.
We then compute three types of metrics: \textbf{(1) pair-wise accuracy:} For each pair consisting of a ground-truth action and a deviation action within the same state, if the ground-truth action receives a score greater than or equal to the deviation action, it is counted as correct; otherwise, it is incorrect. Pair-wise accuracy is the proportion of correct pairs across the whole dataset. \textbf{(2) state-level accuracy:} Evaluates whether the world model consistently ranks the ground-truth action above all deviations within a given state. A state is considered correct only if the ground-truth action has a score greater than or equal to all its corresponding deviation actions. \textbf{(3) task-level accuracy:} Assesses the correctness of an entire task by ensuring that the world model helps correctly select ground-truth action in every state within the task. A task is considered correct only if all its states are correct according to state-level accuracy criteria.
We use state-level accuracy as the primary metric to select the best checkpoint, which closely aligns with the downstream task forms.

\subsection{Results}
\begin{table}[htb]
    \centering
    \small
    \begin{tabular}{cccc}
    \toprule
       World Model & Pair-wise Accuracy & State-level Accuracy & Task-level Accuracy \\
       \midrule
       GPT-4o-mini & 85.30 & 78.01 & 47.73 \\
       GPT-4o & 87.10 & 80.85 & \textbf{52.27} \\
       Qwen2-VL-7B & 86.38 & 80.85 & 50.00 \\
       Qwen2-VL-72B & 87.10 & 80.14 & 47.73 \\
       Dreamer-7B & \textbf{88.53} & \textbf{82.98} & \textbf{52.27} \\
    \bottomrule
    \end{tabular}
    \caption{Results (\%) of various world models on the intrinsic evaluation set.}
    \label{tab:intrinsic-eval}
\end{table}

\begin{figure}
\centering
\includegraphics[width=0.7\linewidth]{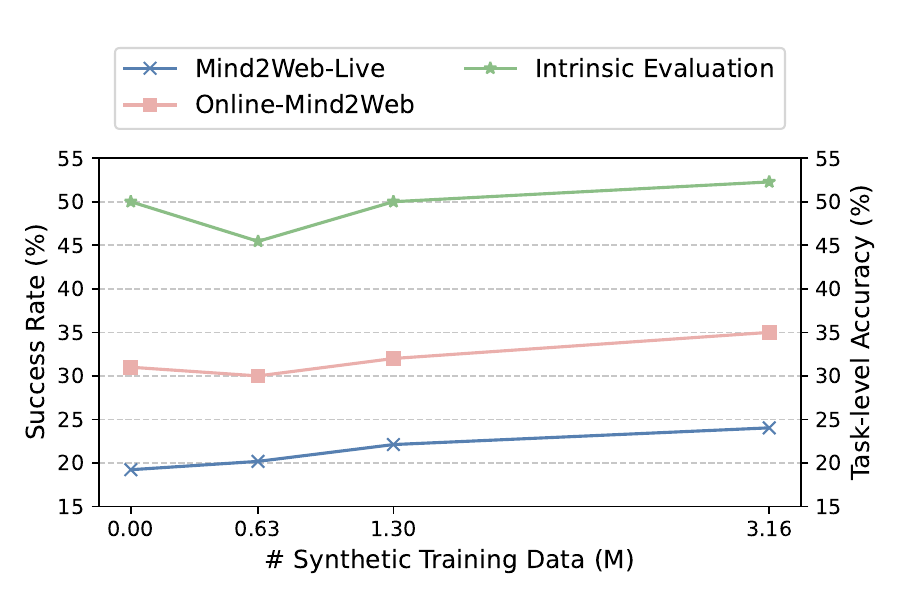}
\caption{Performance on two downstream benchmarks and intrinsic evaluation w.r.t. different sizes of training data.}
\label{fig:intrinsic-scaling}
\end{figure}

Table~\ref{tab:intrinsic-eval} shows the results of the intrinsic evaluation. Our fine-tuned 7B model achieves comparable performance on task-level accuracy with GPT-4o and even outperforms GPT-4o in terms of pair-wise and state-level accuracy. The scaling trend of training world models can also be observed in intrinsic evaluation as shown in Figure~\ref{fig:intrinsic-scaling}.


For all world models we have evaluated on both intrinsic evaluation and Mind2Web-Live~\citep{webcanvas}, the Pearson correlation coefficient between task-level accuracy and task success rate is 0.8455, indicating a strong correlation between intrinsic evaluation and downstream performance. We hope our intrinsic evaluation can serve as a useful tool for advancing web world model development.


\section{Case Study}
\label{appendix:case-study}

\subsection{Error Caused by Imperfect World Model Simulation}
\label{appendix:wm-error}
\begin{figure*}[h]
    \centering
    \includegraphics[width=1\linewidth]{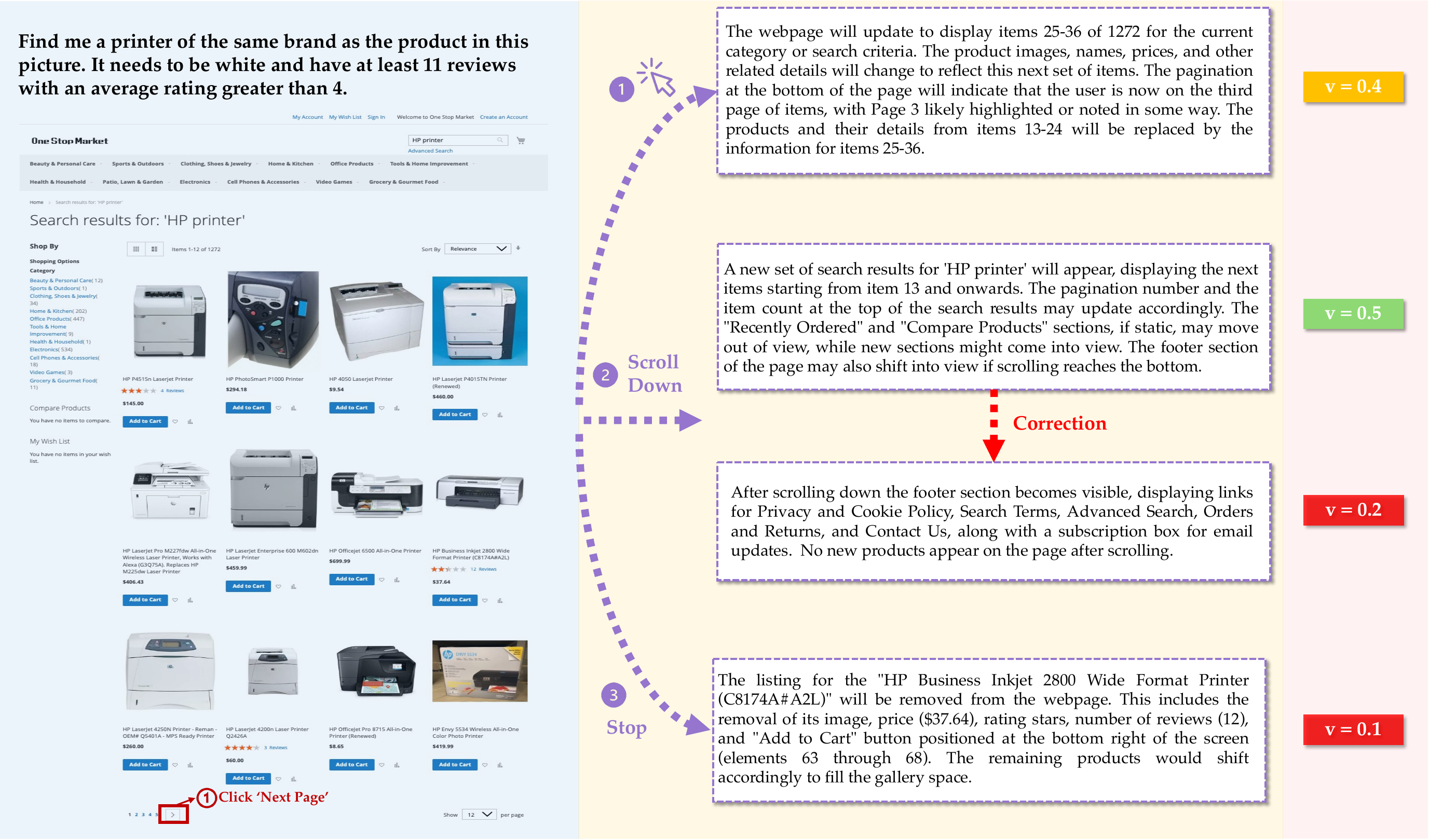}
    \caption{An error case caused by imperfect world model simulation. }
    \label{fig:wm_error}
\end{figure*}

An error case caused by imperfect world model simulation is shown in Figure~\ref{fig:wm_error}.

\subsection{Positive Case Benefiting from World Model Simulation}
\label{appendix:postive-case-1}
\begin{figure*}[h]
    \centering
    \includegraphics[width=1\linewidth]{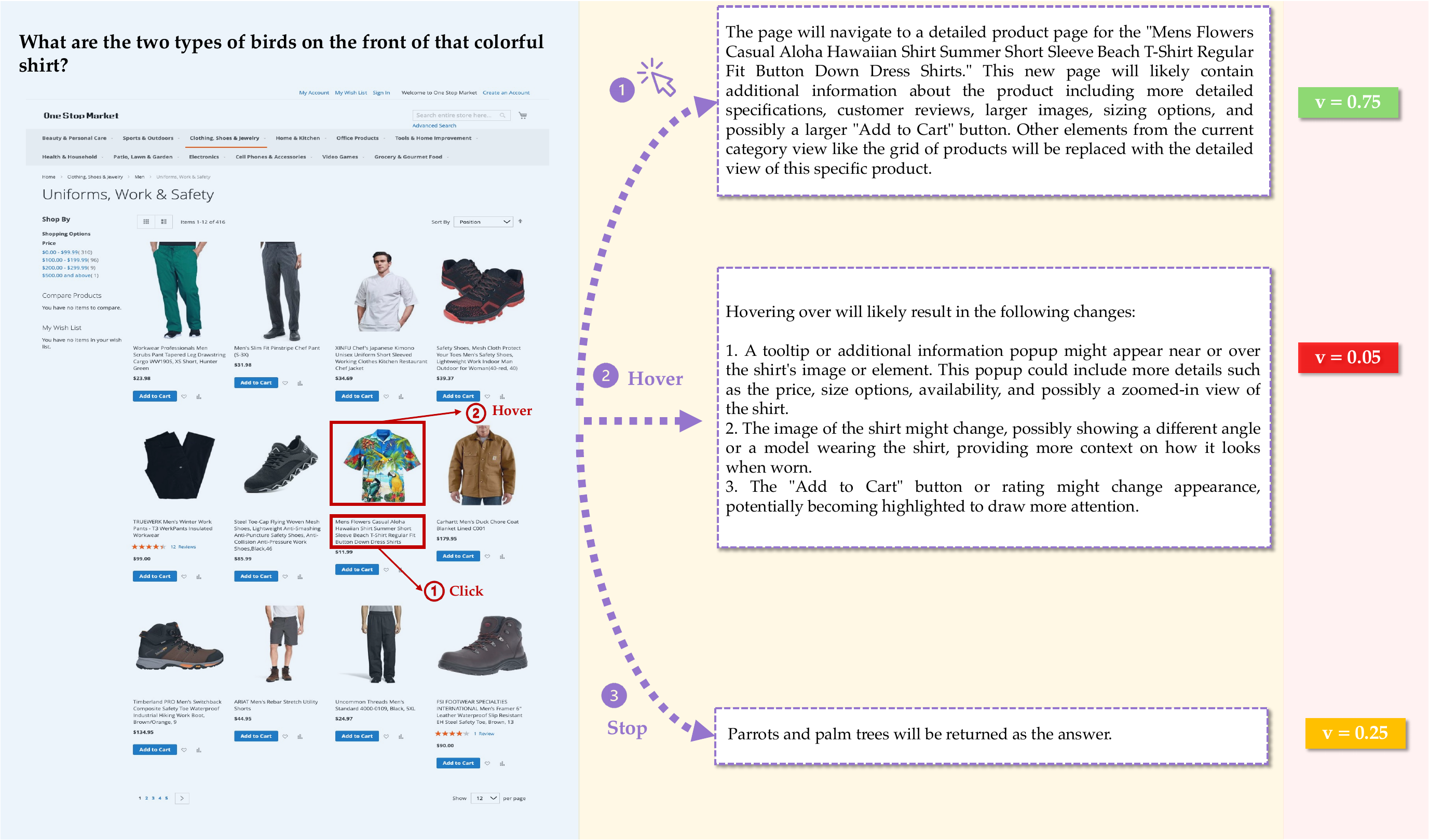}
    \caption{A positive case where the simulation leads to correct action prediction.}
    \label{fig:wm_correct}
\end{figure*}

A positive case where the simulation leads to correct action prediction is shown in Figure~\ref{fig:wm_correct}.

\end{document}